\documentclass[11pt,a4paper]{article}
\usepackage{authblk}
\usepackage[hyperref]{acl2021}
\usepackage{times}
\usepackage{latexsym}

\usepackage{comment}
\usepackage{graphicx}
\usepackage{amsmath}
\usepackage{booktabs}
\usepackage{multirow}
\usepackage{hyperref}
\usepackage{todonotes}

\usepackage{microtype}

\aclfinalcopy 

\title{Domain-independent User Simulation  with Transformers for Task-oriented Dialogue Systems}

\date{}

\begin{document}
\author[1]{Hsien-chin Lin}
\author[1]{Nurul Lubis}
\author[2]{Songbo Hu}
\author[1]{Carel van Niekerk}
\author[1]{\\Christian Geishauser}
\author[1]{Michael Heck}
\author[1]{Shutong Feng}
\author[1]{Milica {Gašić}}
\affil[1]{Heinrich Heine University Dusseldorf, Germany}
\affil[2]{Department of Computer Science and Technology, University of Cambridge, UK}
\affil[1]{\texttt{\{linh,lubis,niekerk,geishaus,heckmi,shutong.feng,gasic\}@hhu.de}}
\affil[2]{\texttt{sh2091@cam.ac.uk}}

\maketitle
\begin{abstract}
Dialogue policy optimisation via reinforcement learning requires a large number of training interactions, which makes learning with real users time consuming and expensive. Many set-ups therefore rely on a user simulator instead of humans. These user simulators have their own problems. While hand-coded, rule-based user simulators have been shown to be sufficient in small, simple domains, for complex domains the number of rules quickly becomes intractable. State-of-the-art data-driven user simulators, on the other hand, are still domain-dependent. This means that adaptation to each new domain requires redesigning and retraining. In this work, we propose a domain-independent transformer-based user simulator (TUS). The structure of our TUS is not tied to a specific domain, enabling domain generalisation and learning of cross-domain user behaviour from data. We compare TUS with the state of the art using automatic as well as human evaluations. TUS can compete with rule-based user simulators on pre-defined domains and is able to generalise to unseen domains in a zero-shot fashion.

\end{abstract}

\section{Introduction}
Task-oriented dialogue systems are designed to help users accomplish specific goals within a particular task such as hotel booking or finding a flight. 
Solving this problem typically requires tracking and planning \cite{young2002talking}. 
In tracking, the system keeps track of information about the user goal from the beginning of the dialogue until the current dialogue turn. 
In planning, the dialogue policy makes decisions at each turn to maximise future rewards at the end of the dialogue \cite{levin1997stochastic}. 
The system typically needs thousands of interactions to train a usable policy \cite{schatzmann-etal-2007-agenda, pietquin2011sample, li2016user, shi2019build}.
The amount of interactions required makes learning from real users time-consuming and costly.
It is therefore appealing to automatically generate a large number of dialogues with a user simulator (US)
\footnote{There are approaches that attempt to learn a dialogue policy from direct interaction with humans~\cite{gavsic2011line}. Even then, USs are essential for development and evaluation.}\cite{eckert1997user}.

Rule-based USs are interpretable and have shown success when applied in small, simple domains.
However, expert knowledge is required to design their rules and the number of rules needed for complex domains quickly becomes intractable~\cite{schatzmann-etal-2007-agenda}.
In addition, handcrafted rules are unable to capture human behaviour to its fullest extent, leading to suboptimal performance when interacting with real users~\cite{schatzmann2006survey}.

Data-driven USs on the other hand can learn user behaviour directly from a corpus. However, they are still domain-dependent. This means that in order to accommodate an unseen domain one needs to collect and annotate a new dataset, and retrain or even re-engineer the simulator. 

We propose a transformer-based domain-independent user simulator (TUS). Unlike existing data-driven simulators, we design the feature representation to be domain-independent, allowing the simulator to easily generalise to new domains without modifying or retraining the model. We utilise a transformer architecture \citep{NIPS2017_3f5ee243} so that the input sequence can have a variable length and dynamic order. The dynamic order takes into account the user's priorities and the varying input length enables the US to incorporate system actions in a seamless manner.
TUS predicts the value of each slot and the domains of the current turn, allowing the model to optimise its performance in multiple granularities.
By disentangling the user behaviour from the domains, TUS can learn a more general user policy to train the dialogue policy. 

We compare policies trained with our TUS to policies trained with other USs through indirect and direct evaluation as well as human evaluation.
The results show that policies trained with TUS outperform those that are trained with another data-driven US and are on par with policies trained with the agenda-based US (ABUS). Moreover, the policy generalises better when evaluated with a different US.
Automatic and human evaluations on our zero-shot study show that leave-one-domain-out TUS is able to generalise to unseen domains while maintaining a comparable performance to ABUS and TUS trained on the full training data.

\section{Related Work}
The quality of a US has a significant impact on the performance of a reinforcement-learning based task-oriented dialogue system~\cite{schatzmann2005quantitative}. 
One of the early models include an N-gram user simulator proposed by \newcite{eckert1997user}. 
It uses a 2-gram model $P(a_u|a_m)$ to predict the user action $a_u$ according to the system action $a_m$. 
Since it only has access to the latest system action, its behaviour can be illogical if the goal changes. 
Therefore, models which can take into account a given user goal were introduced \cite{georgila2006user, eshky-etal-2012-generative}.
The Bayesian model of \newcite{daubigney2012comprehensive} predicts the user action based on the user goal, and hidden Markov models are used to model the user and the system behaviour \cite{cuayahuitl2005human}. 
The graph-based US of \newcite{scheffler2002automatic} combines all possible dialogue paths in a graph. 
It can generate reasonable and consistent behaviour, but is impractical to implement, since extensive domain knowledge is required.

The agenda-based user simulator (ABUS) \cite{schatzmann-etal-2007-agenda} models the user state as a stack-like agenda, ordered according to the priority of the user actions. 
The probabilities of updating the agenda and choosing user actions are set manually or learned from data~\cite{keizer2010parameter}. 
Still, the stacking and popping rules are domain-dependent and need to be designed carefully.

To build a data-driven model, the sequence-to-sequence (Seq2Seq) model structure is widely used. 
\newcite{el2016sequence} propose a Seq2Seq semantic level US with an encoder-decoder structure.
Each turn is fed into the encoder recurrent neural network (RNN) and embedded as a context vector.
Then this context vector is passed to the decoder RNN to generate user actions.
To add new domains, it is necessary to modify the domain-dependent feature representation and retrain the model.

Instead of generating semantic level output, the neural user simulator (NUS) by \newcite{kreyssig-etal-2018-neural} generates responses in natural language, thus requiring less labeling, at the expense of interpretability. However, its feature representation is still domain-dependent.

A variational hierarchical Seq2Seq user simulator (VHUS) is proposed by \newcite{gur2018user}.
Instead of designing dialogue history features, the model encodes the user goal and system actions with a vector using an RNN, which alleviates the need of heavy feature engineering.
However, the inputs are represented as one-hot encodings, which are also dependent on the ontology.
In addition, the output generator is not constrained by the ontology in any way, so it can generate impossible actions.

\begin{figure}[t]
    \centering
    \includegraphics[width=6cm]{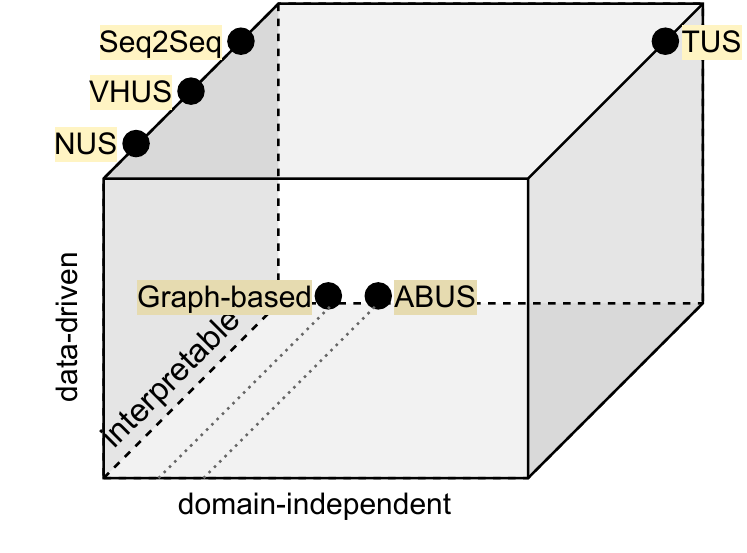}
    \caption{The difference between USs. We compare to which extent a model is data-driven, domain-independent and interpretable.}
    \label{fig:box}
\end{figure}

As shown in Fig.~\ref{fig:box}, ABUS and graph-based models are domain-dependent and require significant design efforts.
Data-driven models such as Seq2Seq, NUS, and VHUS can learn from data, but are constrained by the underlying domain. NUS generates natural language responses, which requires less labeling, but comes with reduced interpretability. 

\newcite{shi2019build} compared different ways to build a US and indicated that the data-driven models suffer from bias in the corpus.
If some actions are rare in the corpus, the model cannot capture them.
Thus, the dialogue policy cannot explore all possible paths during training with the data-driven USs.
It is important to learn more general human behaviour to reduce the impact of the corpus bias.


\section{Problem Description}
\begin{figure}[t]
    \centering
    \includegraphics[width=7.7cm]{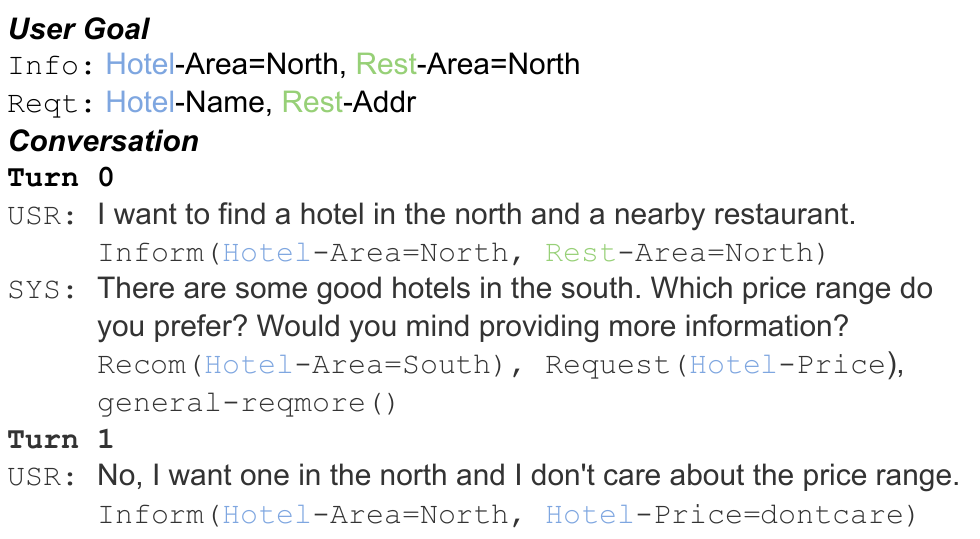}
    \caption{An example dialogue with a multi-domain goal.}
    \label{fig:task-ex}
\end{figure}

Task-oriented dialogue systems are defined by a given \emph{ontology}, which specifies the concepts that the system can handle. 
The ontology can include multiple \emph{domains}.
In each domain, there are \emph{informable slots}, which are the attributes that users can assign \emph{values} to, and \emph{requestable slots}, which are the attributes that users can query.
For example, in Fig.~\ref{fig:task-ex} the user goal has two domains, ``hotel'' and ``restaurant''. The slot \texttt{Area} is an informable slot with the value \texttt{North} in domain ``hotel'' and \texttt{Addr} is a requestable slot in domain ``restaurant''. The \emph{system state} records the slots and values mentioned in the dialogue history.
A US for task-oriented dialogue systems needs to provide coherent responses according to a given user goal $G = \{ domain_1 : [ (slot_1, value_1), (slot_2, value_2), \dots], \dots\}$. The domains, slots and values are selected from the ontology.

The \emph{user action} is composed of user intents, domains, slots, and values. We consider user intents that appear in the MultiWOZ dataset~\cite{budzianowski-etal-2018-multiwoz}. It is of course possible to consider arbitrary intents within the same model architecture, as long as they are defined a priori\footnote{We note that intents are not normally dependent on the domain but rather on the kind of dialogue that is being modeled, e.g. task-oriented or chit-chat.}. 
The two possible user intents we consider are \emph{Inform} and \emph{Request}. With \emph{Inform}, the user can provide information, correct the system or confirm the system's recommendations.
When a user goal cannot be fulfilled, the user can also randomly select a value from the ontology and change the goal. With \emph{Request}, the user can request information about certain slots.

The \emph{system action} is similar to the user action, but there exist more (system) intents. For example, the system can provide suggestions to users with the intent \emph{Recommendation} and make reservations for users with the intent \emph{Book}. More system intents can be found in Appendix~\ref{ap:all_sys_intent}.

We view user simulation in a task-oriented dialogue as a sequence-to-sequence problem.
For each turn $t$, we extract the input feature vectors $V^t$ of the input list of slots $S^t = [s_1, s_2, \dots]$, which is composed of the slots from the user goal and the system action. The output sequence $O^t = [o_1^t, o_2^t, \dots]$ is then generated by the model, where $o_i^t$ shows how the value for slot $s_i$ is obtained. The input feature representation and the output target should be domain-independent in order to generalise to unseen domains without redesigning and retraining. More details can be found in Sec.~\ref{sec:model_detail}.

By working on the semantic level during training, we retain interpretability. To interact with real users during human evaluation, we rely on template-based natural language generation to convert the semantic-level actions into utterances, as language generation is out of the scope of this work.

\section{Transformer-based Domain-independent User Simulator}
\label{sec:model_detail}
\begin{figure*}[h]
    \centering
    \includegraphics[width=14cm]{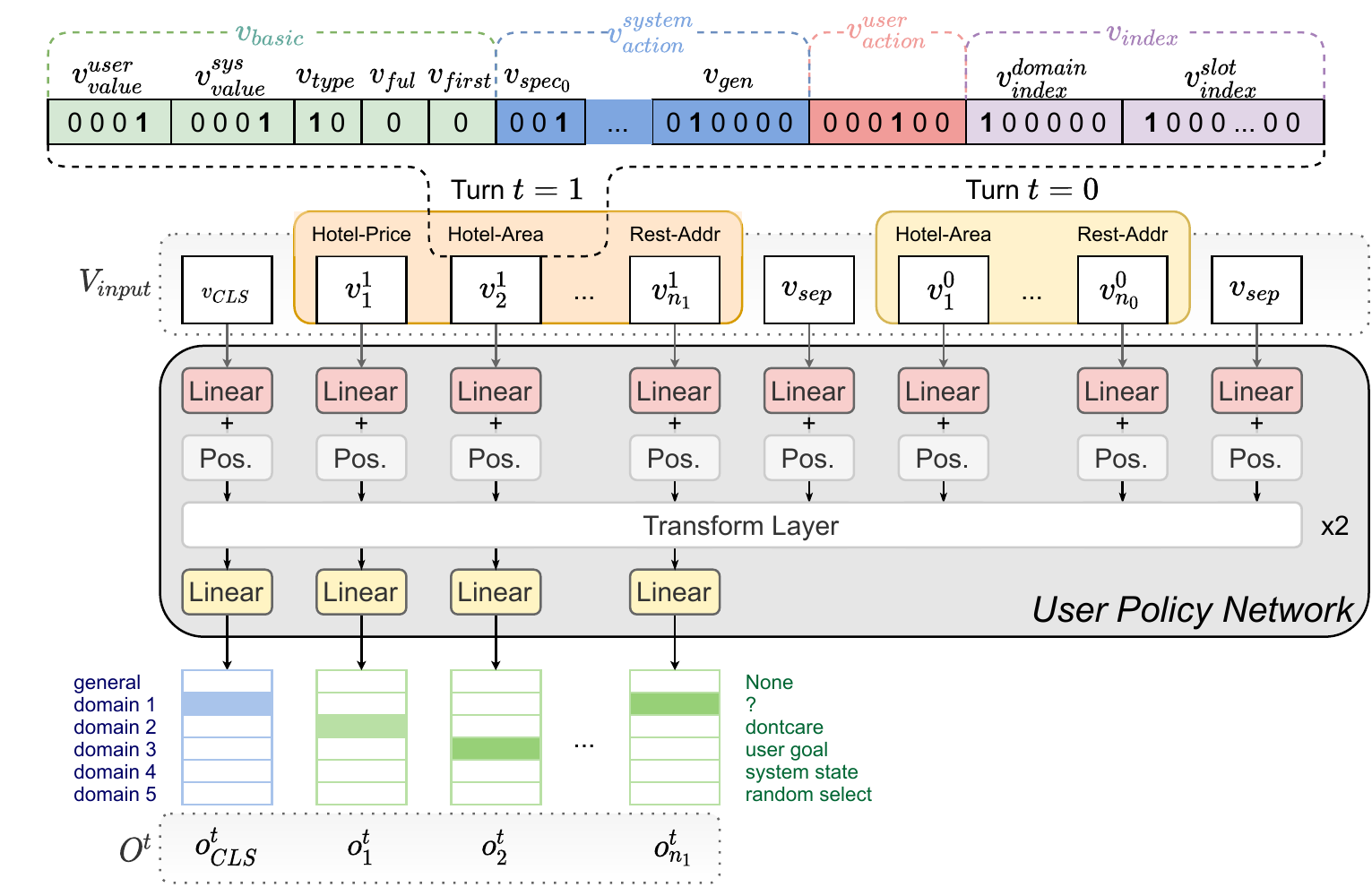}
    \caption{The TUS model structure. The input list starts with a special token, \texttt{[CLS]}, and comprises slot lists from previous turns. The slot lists from each turn are separated by a token, \texttt{[SEP]}.  The model predicts an output vector for each slot in the last turn. Note that the order of slots in each turn is independent from each other. The output for \texttt{[CLS]} represents which domains should be selected in the current turn. The user goal and dialogue history are shown in Fig.~\ref{fig:task-ex} and here we give the example of the input feature $v_i$ for slot \texttt{Hotel-Area}. }
\label{fig:TUS}
\end{figure*}

The TUS model structure is shown in Fig.~\ref{fig:TUS}.
For each turn $t$, the list of input feature vectors $V^t = [v^t_1, v^t_2 \dots, v^t_{n_t}]$ is generated based on the system actions and the user goal, where $v^t_i$ is the feature vector of slot $s_i$ and $n_t$ is the length of the input list in turn $t$, $V^t$. 
We explain the feature representation in detail in Sec.~\ref{sec:feature}.
Inspired by ABUS, which models the user state as a stack-like agenda, the length of input list $n_t$ at each turn $t$ varies by taking into account slots mentioned in the system's action.
For example, in Fig.~\ref{fig:TUS} the input list $V^0$ only contains the slots in the user goal at the first turn.
Then the system mentions a slot not in the user goal, \texttt{Hotel-Price}.  So in turn $1$ the length of input list $V^1$ is $n_1 = n_0 + 1$ because one slot is inserted into the input list $V^1$.
The whole input sequence to the model is $V_{input} = [v_{CLS}, v^t_1, \dots, v_{SEP}, v^{t-1}_1, \dots, v_{SEP}]$, where $v_{CLS}$ is the representation of \texttt{[CLS]} and $v_{SEP}$ is the representation of \texttt{[SEP]}.

The user policy network is a transformer \cite{NIPS2017_3f5ee243, devlin-etal-2019-bert}. 
We choose this structure because transformers are able to handle input sequences of arbitrary lengths and to capture the relationship between slots thanks to self-attention.
The model structure includes a linear layer and position encoding for inputs, two transformer layers, and one linear layer for outputs.

The output list $O^t = [o_1^t, \dots, o_{n_t}^t]$ consists of one-hot vectors $o_i^t$ which determine the values of the slots $s_i$ at turn $t$. 
The dimensions of $o_i^t \in \{0,1\}^6$ correspond to ``none'', ``don't care'', ``?'', ``from the user goal'', ``from the system state'', or ``randomly selected''.
More precisely, ``none'' means that this slot is not mentioned in this turn, ``don't care'' signifies that the US does not care about this slot, ``?'' means the US wants to request information about this slot, ``from user goal'' implies that the value is the same as in the user goal, ``from system state'' means that the value is as mentioned by the system, and lastly ``randomly selected'' indicates that the US wants to change its goal by randomly selecting a value from the ontology.

The loss function for slots measures the difference between the predicted output $O^t$ and the target $Y^t$ at each turn $t$ from the dataset as computed by cross entropy (CE), i.e.,
\begin{equation}
    loss_{slots} = \frac{1}{n_t}\sum_{i=1}^{n_t}\text{CE}(o^t_i, y^t_i),
\end{equation}
where $n_t$ is the number of slots in the input list, $o^t_i$ is the output, and $y^t_i$ is the target of slot $s_i$ in turn $t$.

\subsection{Domain-independent Input Features}
\label{sec:feature}

We design the input feature representation $v_i^t$ of each slot $s_i$ in turn $t$ consisting of a set of sub-vectors, all of which are domain-independent.
For better readability, we drop the slot index $i$ and the turn index $t$, i.e.\ we write $v$ for $v_i^t$.

\subsubsection{Basic Information Features}
Inspired by the feature representation proposed in \newcite{el2016sequence},  we use a feature vector $v_{basic}$ that is composed of binary sub-vectors to represent the basic information for each slot.
Each slot has two value vectors: $v^{sys}_{value}$ represents the value in the system state, and $v^{user}_{value}$ represents the value in the user goal.
Each value vector is a 4-dimensional one-hot vector,
 with coordinates encoding ``none'', ``?'', ``don't care'' or ``other values'', in this order.
For example, in turn 1 in Fig.~\ref{fig:task-ex}, for slot \texttt{Hotel-Price} $v^{user}_{value} = [1, 0, 0, 0]$, i.e., ``none'',  because it is not in the user goal,  and $v^{sys}_{value} = [0, 1, 0, 0]$, i.e., ``?'', because the system requests it.

The slot type vector $v_{type}$ is a 2-dimensional vector which represents whether a slot is in the user goal as a constraint or a request. For example, in Fig.~\ref{fig:task-ex} for \texttt{Hotel-Area} $v_{type} = [1, 0]$ (constraint), while for \texttt{Hotel-Name} $v_{type} = [0, 1]$ (request). A value of $[0, 0]$ means that the slot is not included in the user goal.

The state vector $v_{ful}$ encodes whether or not a constraint or informable slot has been fulfilled. The value is set to $1$ if the constraint has been fulfilled, and to $0$ otherwise.
The vector $v_{first}$ similarly encodes whether a slot is mentioned for the first time. 

The basic information feature vector $v_{basic}$ is the concatenation of these vectors, i.e.,
\begin{equation}
    v_{basic} = v^{user}_{value} \oplus v^{sys}_{value} \oplus  v_{type}\oplus  v_{ful} \oplus v_{first}
\end{equation}

\subsubsection{System Action Features}
The system action feature vector $v_{action}^{system}$ encodes system actions in each turn.  
There are two kinds of system actions, general actions and domain-specific actions. 
The general actions are composed only with general intents, such as ``reqmore'' and ``bye''. 
For example, \texttt{general-reqmore()}. 
The feature vector of general actions $v_{gen}$ is a multi-hot encoding of whether or not a general intent appears in the dialogue. 
With a total number of $n_{gen}$ general intents, for each $k \in \{1, \dots, n_{gen}\}$, the $k$-th entry of $v_{gen}$ is set to 1 if the $k$-th general intent is part of the system act.

On the other hand, domain-specific actions are composed with domains, slots, values, and domain-specific intents such as ``recommend'' and ``select''.
For example, \texttt{Recom(Hotel-Area=South)}. 
Each domain-specific action vector $v_{spec_j}$ with the domain-specific $j$-th intent, $j \in \{1, \dots, n_{spec}\}$, where $n_{spec}$ is the total number of domain-specific intents, is represented by a 3-dimensional one-hot encoding that describes whether the value is ``none'', ``?'' or ``other values''.

The final action representation $v_{action}^{system}$ is formed by concatenating $n_{spec}$ domain-specific action representations together with the general action representation, i.e.,
\begin{equation}
    v_{action}^{system} = v_{spec_0} \oplus \dots \oplus v_{spec_{n_{spec}}} \oplus v_{gen}.
\end{equation}
For the slot \texttt{Hotel-Area} in Fig.~\ref{fig:TUS}, we have a vector for each intent. For the intent ``recommend'' $v_{spec_0} = [0, 0, 1]$, which means that ``other values'' (in this case \texttt{South}) are mentioned. For all other domain-specific intents, the vectors are $[1, 0, 0]$ since no value is mentioned. In terms of the general intents, only ``reqmore'' is mentioned, so $v_{gen}[1] = 1$, as ``reqmore'' is the first general intent. 

\subsubsection{User Action Features}
The output vector from the previous turn $O^{t-1}$ is also included in the input features of the next turn $t$ to take into account what has been mentioned by the US itself, i.e.\ for slot $s_i$ in turn $t$, the user action feature $v_{action}^{user}=o^{t-1}_i$.

\subsubsection{Domain and Slot Index Features}
In some cases, multiple slots may share the same basic feature $v_{basic}$, system action feature $v_{action}^{system}$ and user action feature $v_{action}^{user}$. 
This similarity in features of different slots makes it difficult for the model to distinguish one slot from another, despite  the positional encoding. 
In particular, it is challenging for the model to learn the relationship between turns for a given slot because the number and the order of slots vary from one turn to the next.
This may lead to over-generation: the model selects all slots with the same feature vector.

To counteract this issue, we introduce the index feature $v_{index}$, which consists of the domain index feature $v^{domain}_{index} \in \{0, 1\}^{l_d}$ and the slot index feature $v^{slot}_{index} \in \{0, 1\}^{l_s}$, where $l_d$ is the maximum number of domains in a user goal and $l_s$ is the maximum number of slots in any given domain\footnote{This does not need to be dependent on the number of domains or slots, it can simply be a random identifier assigned to each slot during one dialogue.}.

To make the index feature ontology-independent, for a particular slot, $v_{index}$ remains consistent throughout a dialogue, but varies between dialogues. The order of the index in each dialogue is determined by the order in the user goal. 
For example, the ``hotel'' domain can be the first domain in one user goal of the first dialogue, and the second domain in the next.

Then for each slot in each turn the input feature vector $v$ is formed by concatenating all sub-vectors:
\begin{equation}
    v = v_{basic} \oplus v_{action}^{system} \oplus v_{action}^{user} \oplus v_{index}.
\end{equation}
An example of $v$ for slot \texttt{Hotel-Area} is shown in Fig.~\ref{fig:TUS} based on the dialogue history in Fig.~\ref{fig:task-ex}. Examples of how the feature representation is constructed can be seen in Appendix~\ref{ap:example}.

\subsection{Domain Prediction}
Inspired by solving downstream tasks using BERT~\cite{devlin-etal-2019-bert}, we utilise the output of \texttt{[CLS]}, $o_{CLS}$, to predict which domains are considered in turn $t$ as a multi-label classification problem. 
The domain loss $loss_{domain}$ measures the difference between the output $o_{CLS}$ and the target $y_{CLS}$ for each turn by binary cross entropy (BCE).
The final loss function is defined as
\begin{equation}
    loss = loss_{slots} + loss_{domain}.
\end{equation}

\section{Experimental Setup}
\subsection{Supervised Training for TUS}
Our model is implemented in PyTorch \cite{NEURIPS2019_9015} and optimised using the Adam optimiser \cite{DBLP:journals/corr/KingmaB14} with learning rate $5\times 10^{-4}$.
The dimension of the input linear layer is $100$, the number of the transformer layers is $2$, and the dimension of the output linear layer is $6$.
The maximum number of domains $l_d$ is 6 and the maximum number of slots in one domain $l_s$ is 10.
During training, the dropout rate is $0.1$.

We train our model\footnote{\url{https://gitlab.cs.uni-duesseldorf.de/general/dsml/tus_public}} on the MultiWOZ 2.1 dataset~\cite{eric2020multiwoz}, consisting of dialogues between two humans, one posing as a user and the other as an operator. 
The dialogues in the dataset are complex because there may be more than one domain involved in one dialogue, even in the same turn. 
During training and testing with the dataset, the order of slots in the input list is derived from the data, which means slot $s_i$ is before slot $s_{i+1}$ if the user mentioned slot $s_i$ first.
For inference without the dataset, such as when using TUS to train a dialogue policy, the order of slots is randomly generated.

We measure how well a US can fit the dataset by precision, recall, F1 score, and turn accuracy.
The turn accuracy measures how many model predictions per turn are identical to the corpus, based on the oracle dialogue history.

\subsection{Training Policies with USs}
User simulators are designed to train dialogue systems, thus a better user simulator should result in a better dialogue system.
We train different dialogue policies by proximal policy optimization (PPO) \cite{schulman2017proximal}, a simple and stable reinforcement learning algorithm, with ABUS, VHUS, and TUS as USs in the ConvLab-2 framework \cite{zhu2020convlab2}.
The policies are trained for 200 epochs, each of which consists of 1000 dialogues.
The reward function gives a reward of 80 for a successful dialogue and of -1 for each dialogue turn, with the maximum number of dialogue turns set to 40. For failed dialogues, an additional penalty is set to -40. 
Each dialogue policy is trained on 5 random seeds.
The dialogue policies are then evaluated using all USs by cross-model evaluation \cite{schatztnann2005effects} to demonstrate the generalisation ability of the policy trained with a particular US when evaluated with a different US. 

\subsection{Leave-one-domain-out Training}
\label{sec:leave}

To evaluate the ability of TUS in handling unseen domains, we remove one domain during supervised learning of TUS.
The leave-one-domain-out TUSs are used to train dialogue policies with all possible domains.
For example, TUS-noHotel is trained on the dataset without the ``hotel'' domain.
During policy training, the user goal is generated randomly from all possible domains.

Some domains in MultiWOZ may share the same slots, such as ``restaurant'' and ``hotel'' domains which contain property-related slots, e.g. ``area,'' ``name," and ``price range.'' However, the corpus also includes domains that are quite different from the rest, For example, the ``train" domain which contains many time-related slots such as ``arrival time'' or ``departure time'', as well as unique slots such as ``price" and ``duration." The different properties of the domains will allow us to study the zero-shot transfer capability of the model.

\subsection{Human Evaluation}
Following the setting in \newcite{kreyssig-etal-2018-neural}, we select 2 of the 5 trained versions of each dialogue policy for evaluation 
in a human trial:
the version performing best on ABUS, and the version performing best in interaction with TUS.
The results of the two versions are averaged.
For each version we collect 200 dialogues, which means there are 400 dialogues for each policy in total.  
Dialogue policies trained with VHUS significantly underperform, so we only consider policies trained with ABUS or TUS for the human trial (see Table~\ref{tab:cross-model}).
The best and the worst policies in the leave-one-domain-out experiment are also included to see the upper and lower bound of the zero-shot domain generalisation performance.

Human evaluation is performed via DialCrowd \cite{lee-etal-2018-dialcrowd} connected to Amazon Mechanical Turk\footnote{\url{https://www.mturk.com/}}. 
Users are provided with a randomly generated user goal and are required to interact with our systems in natural language.

\section{Experimental Results}

\subsection{Cross-model Evaluation}
The results of our experiments are shown in Table~\ref{tab:cross-model}. 
The policy trained with TUS performs well when evaluated with ABUS, with 10\% absolute improvement in the success rate over its performance on TUS. On the other hand, while a policy trained with ABUS performs almost perfectly when evaluated with ABUS, the performance drops significantly, by 35\% absolute, when this policy interacts with TUS. This signals that, in the case of ABUS, the policy overfits to the US used for training, and is not able to generalise well to the behaviour of other USs. 
We found that VHUS is neither able to train nor to evaluate a multi-domain policy adequately. This was also observed in the experiments by \newcite{takanobu-etal-2019-guided}. We suspect that this is due to the fact that VHUS was designed to operate on a single domain and does not generalise well to the multi-domain scenario. To the best of our knowledge, no other data-driven US has been developed for the multi-domain scenario.

\begin{table}[t]
\small
\centering
\begin{tabular}{@{}lrrrr@{}}
\toprule
\multicolumn{1}{c}{US for} & \multicolumn{4}{c}{US for evaluation} \\
\multicolumn{1}{c}{training} & \multicolumn{1}{c}{ABUS} & \multicolumn{1}{c}{VHUS} & \multicolumn{1}{c}{TUS} & \multicolumn{1}{c}{avg.} \\ \midrule
ABUS & 0.93 & 0.09 & 0.58 & 0.53\\
VHUS & 0.62 & 0.11 & 0.37 & 0.36 \\
TUS & 0.79 & 0.10 & 0.69 & 0.53 \\ \bottomrule
\end{tabular}
\caption{The success rates of policies trained on ABUS, VHUS, and TUS when tested on various USs.}
\label{tab:cross-model}
\end{table}

\begin{figure}[t]
    \centering
    \includegraphics[width=7.7cm]{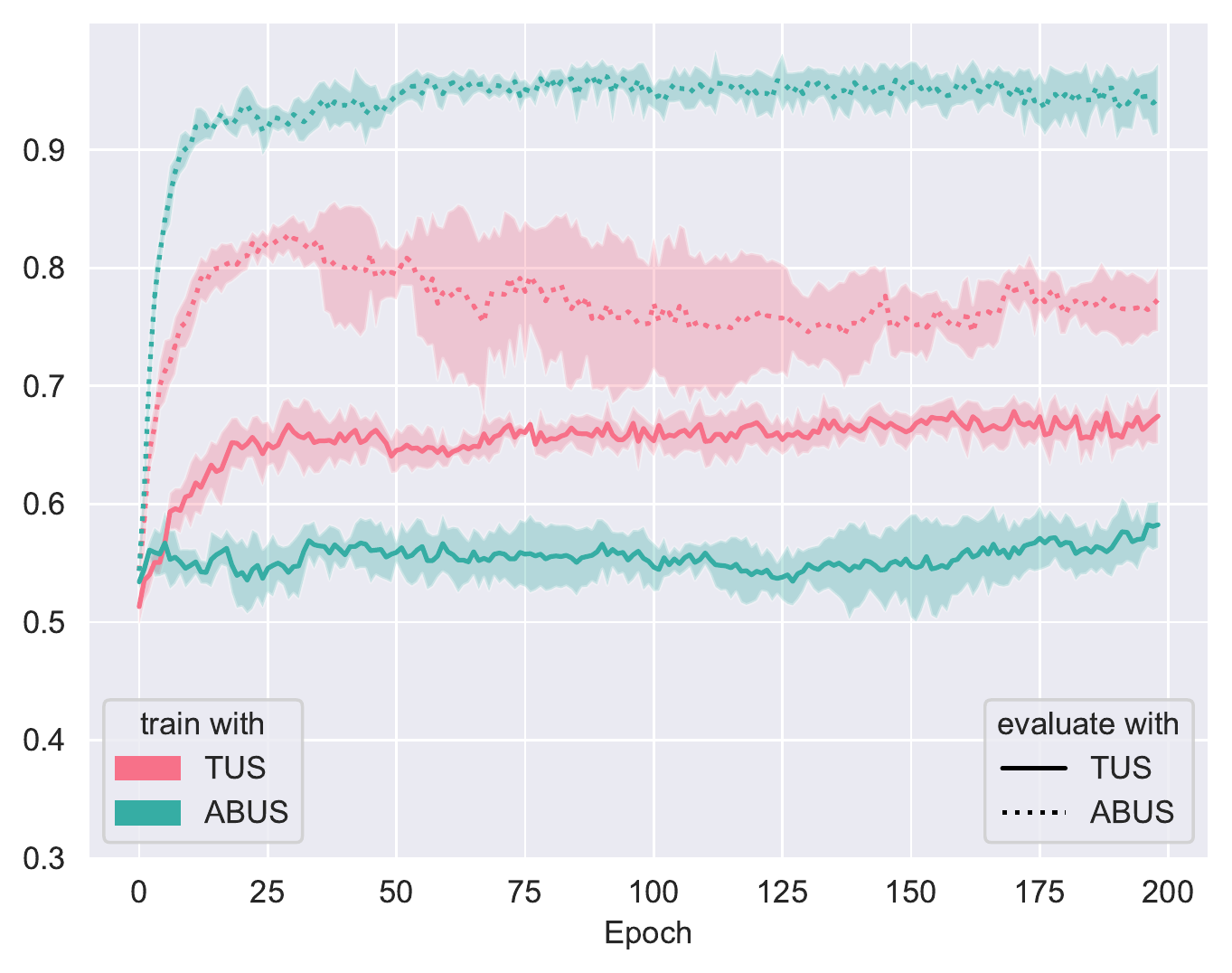}
    \caption{The success rates of policies during training with TUS and ABUS.}
    \label{fig:success}
\end{figure}

The success rates of policies trained with ABUS and TUS during training, evaluated with both US, are shown in Fig.~\ref{fig:success}.
Each of the systems is trained 5 times on different random seeds. We report the average success rate as well as the standard deviation.
Although the policy trained with TUS is more unstable when evaluated on ABUS, it still shows an improvement from the initial policy, converging at around 79\%. On the other hand, the policy trained with ABUS and evaluated with TUS barely show any improvements.

\subsection{Impact of features and loss functions}
We conduct an ablation study to investigate the usefulness of the proposed features and loss functions.
The result is shown in Table.~\ref{tab:ablation}.
First, we measure the performance of the basic model which uses only the basic information feature $v_{basic}$, the system action feature $v_{action}^{system}$, and the user action feature $v_{action}^{user}$ as the input.
While this model can have a high recall rate, the precision and the turn accuracy are fairly low. 
We deduce that without the index features the model cannot distinguish the difference between slots and therefore tends to select slots of the same slot type in one turn. For example, it provides all constraints in the first turn, which leads to high recall and over-generation. 

\begin{table}[t]
\small
\centering
\begin{tabular}{@{}lllllr@{}}
\toprule
method & P & R & F1 & ACC & LEN \\ \midrule
basic model & 0.11 & 0.71 & 0.19 & 0.11 & 4.51 \\
\quad+ index feature & 0.17 & 0.51 & 0.26 & 0.44 & 1.29 \\
\quad\quad+ domain loss & 0.17 & 0.54 & 0.26 & 0.46 & 1.22 \\ \bottomrule
\end{tabular}
\caption{The TUS ablation experiments. We analyse the impact of different settings by measuring precision P, recall R, F1 score, turn accuracy ACC, and the average slots mentioned in the first turn user action LEN. Humans, on average, mention 1.5 slots in the first turn.}
\label{tab:ablation}
\end{table}

\begin{table*}[h]
\small
\centering
\setlength{\tabcolsep}{3.75pt}
\begin{tabular}{@{}lrrrrrrrlrrrrrrr@{}}
\toprule
US for & \multicolumn{1}{l}{removed} & \multicolumn{6}{c}{ABUS} &  & \multicolumn{6}{c}{TUS} & \multicolumn{1}{l}{\multirow{2}{*}{mean}} \\ \cmidrule(lr){3-8} \cmidrule(lr){10-15}
training & \multicolumn{1}{l}{data(\%)} & Attr. & Hotel & Rest. & Taxi & Train & all &  & Attr. & Hotel & Rest. & Taxi & Train & all & \multicolumn{1}{l}{} \\ \midrule
TUS-noAttr & 32.20 & \textbf{0.69} & 0.64 & \textbf{0.81} & 0.65 & \textbf{0.75} & \textbf{0.77} &  & 0.71 & 0.58 & 0.66 & 0.61 & \textbf{0.69} & 0.69 & \textbf{0.73} \\
TUS-noTaxi & 19.60 & 0.63 & 0.61 & 0.81 & 0.61 & 0.70 & 0.74 &  & 0.69 & 0.60 & \textbf{0.69} & 0.64 & 0.68 & \textbf{0.69} & 0.72 \\
TUS-noRest & 45.21 & 0.62 & \textbf{0.66} & 0.80 & 0.56 & 0.75 & 0.76 &  & \textbf{0.71} & \textbf{0.60} & 0.64 & \textbf{0.65} & 0.64 & 0.68 & 0.72 \\
TUS-noTrain & 36.95 & 0.64 & 0.65 & 0.78 & \textbf{0.67} & 0.62 & 0.73 &  & 0.67 & 0.54 & \textbf{0.63} & 0.64 & 0.58 & \textbf{0.64} & 0.68 \\
TUS-noHotel & 40.15 & 0.59 & 0.59 & 0.76 & \textbf{0.61} & 0.54 & 0.69 &  & 0.64 & 0.52 & 0.61 & 0.61 & 0.55 & 0.62 & 0.66 \\ \midrule
TUS & 0 & 0.69 & 0.68 & 0.81 & 0.66 & 0.77 & 0.79 &  & 0.73 & 0.59 & 0.66 & 0.68 & 0.64 & 0.69 & 0.74 \\ \bottomrule
\end{tabular}
\caption{The success rates of dialogue policies trained with leave-one-domain-out TUSs. For example, the TUS-noAttr model is trained without the ``attraction'' domain. The sum of all removed data is more than 100\% because some dialogues have multiple domains. We report results on all domains.}
\label{tab:domain}
\end{table*}

Analysis of the generated user actions shows that the basic model tends to mention four or more slots in the first turn. This is unnatural, since human users tend to only mention one or two slots at the beginning of a dialogue. More details about the average slots per turn can be found in Appendix~\ref{ap:average_length}.

After adding the index feature $v_{index}$, the recall rate is decreased by 17\% absolute, but the turn accuracy is increased by 35\% absolute, along with improvements on the precision and the F1 score. Furthermore, the average number of slots per turn is closer to that of a real user. Although the recall rate with respect to the target in the data is decreased, this is not necessarily a concern since in dialogue there are many different plausible actions for a given context.
For example, when searching for a restaurant, we may provide the information of the area first, or the food type. The order of communicating these constraints may vary.

When we include the domain loss $loss_{domain}$ during training, both the recall rate and the turn accuracy improve while a similar average slot length per turn is maintained. These results indicate that the proposed ontology-independent index features can help the model to distinguish one slot from the other, which solves the over-generation problem of the basic model. The domain loss allows for more accurate prediction of the domain at turn level and the value for each slot at the same time.

\subsection{Zero-shot Transfer}

We test the capability of the model to handle unseen domains in a zero-shot experiment. In a leave-one-domain-out fashion we remove dialogues involving one particular domain when training the US. The share of each domain in the total dialogue data ranges from 19.60\% to 45.21\%. During dialogue policy training we sample the user goal from all domains.
As presented in Table~\ref{tab:domain}, removing one domain from the training data when training the US does not dramatically influence the policy on the corresponding domain. The final performance of the policies trained with leave-one-domain-out TUSs is still reasonably comparable to the policy trained with the full TUS. This is especially noteworthy considering the substantial amount of data removed during US training and the difference between each domain. 

We observe that the model is able to learn about the removed domain from the other domains, although the removed domain is different from the remaining ones. For example, the ``train'' domain is very different from ``attraction'', ``restaurant'', and ``hotel'', and it is more complex than ``taxi'', but TUS-noTrain still performs reasonably well on the ``train'' domain. This signals that the model can do zero-shot transfer by leveraging other domain information. The worst performance on the ``train" domain happens instead when the ``hotel" domain is removed, i.e. the domain with the most substantial amount of data.

Our results also show that that some domains are more sensitive to data removal than others, irrespective of which domain is removed. This indicates that some domains are more involved and simply require more training data.
This result demonstrates that TUS has the capability to handle new unseen domains without modifying the feature representation or retraining the model. It also shows that our model is sample-efficient.

\subsection{Human Evaluation}
\begin{table}[t]
\centering
\begin{tabular}{@{}lrrrr@{}}
\toprule
US for & \multicolumn{3}{c}{success} & \multirow{2}{*}{overall} \\ \cmidrule(lr){2-4}
training & Attr. & Hotel & all &  \\ \midrule
ABUS & 0.76 & 0.70 & 0.83 & 3.90 \\
TUS & 0.73 & 0.69 & 0.83 & 4.03 \\
TUS-noAttr & 0.75 & 0.54 & 0.81 & 4.01 \\
TUS-noHotel & 0.73 & 0.55 & 0.76 & 3.86 \\ \bottomrule
\end{tabular}
\caption{The human evaluation results include success rate and overall rating as judged by users.}
\label{tab:human}
\end{table}
The result of the human evaluation is shown in Table~\ref{tab:human}. In total, 156 users participated in the human evaluation. The number of interactions per user ranges from 10 to 80.
The success rate measures whether the given goal is fulfilled by the system and the overall rating grades the system's performance from 1 star (poor) to 5 stars (excellent).
TUS is able to achieve a comparable success rate as ABUS, without domain-specific information, and even scores slightly better in terms of overall rating. We were not able to observe any statistically significant differences between ABUS and TUS in the human evaluation.
For leave-one-domain-out models, the performance of TUS-noAttr is similar to that one of ABUS and TUS without a statistically significant difference.
We do however observe a statistically significant decrease in the success rate of TUS-noHotel when compared to TUS and ABUS $(p < 0.05)$. This is unsurprising as the hotel domain accounts for 40.15\% of the training data.
For both TUS-noAttr and TUS-noHotel, the success rate on the domain ``attraction'' is comparable to TUS and ABUS, but the success rate on the domain ``hotel'' is relatively low. 
As observed in the simulation, removing a domain does not decrease the success rate in the corresponding domain as the feature representation is domain agnostic. Instead, it impacts domains which need plenty of data to learn.

\section{Conclusion}
We propose a domain-independent user simulator with transformers, TUS. We design ontology-independent input and output feature representations.
TUS outperforms the data-driven VHUS and it has a comparable performance to the rule-based ABUS in cross-model evaluation. Human evaluation confirms that TUS can compete with ABUS even though ABUS is based on carefully designed domain-dependent rules. Our ablation study shows that the proposed features and loss functions are essential to model natural user behavior from data.
Lastly, our zero-shot study shows that TUS can handle new domains without feature modification or model retraining, even with substantially fewer training samples.

In future work, we would like to learn the order of slots and add output language generation to make the behaviour of TUS more human-like. Applying reinforcement learning to this model would also be of interest.

\section*{Acknowledgments}

We would like to thank Ting-Rui Chiang and Dr. Maxine Eskenazi from Carnegie Mellon University for their help with the human trial. This work is a part of DYMO project which has received funding from the European Research Council (ERC) provided under the Horizon 2020 research and innovation programme (Grant agreement No. STG2018 804636). N. Lubis, C. van Niekerk, M. Heck and S. Feng are funded by an Alexander von Humboldt Sofja Kovalevskaja Award endowed by the German Federal Ministry of Education and Research. Computational infrastructure and support were provided by the Centre for Information and Media Technology at Heinrich Heine University Düsseldorf. Computing resources were provided by Google Cloud.

\bibliographystyle{acl_natbib}
\bibliography{acl2021}

\appendix

\section{All System Intents}
\label{ap:all_sys_intent}
All  system intents in the MultiWOZ 2.1 dataset are listed in Table~\ref{tab:system_intents}, including 5 general intents and 9 domain-specific intents.
\begin{table}[h]
\centering
\begin{tabular}{p{1cm}p{6cm}}
\toprule
type & intents \\ \midrule
general & welcome, reqmore, bye, thank, greet \\ \midrule
domain-specific & recommend, inform, request,  select, book, nobook, offerbook, offerbooked, nooffer \\ \bottomrule
\end{tabular}
\caption{All system intents in the MultiWOZ 2.1}
\label{tab:system_intents}
\end{table}

\section{Average Action Length in Each Turn}
\label{ap:average_length}
\begin{figure}[h]
    \centering
    \includegraphics[width=7.7cm]{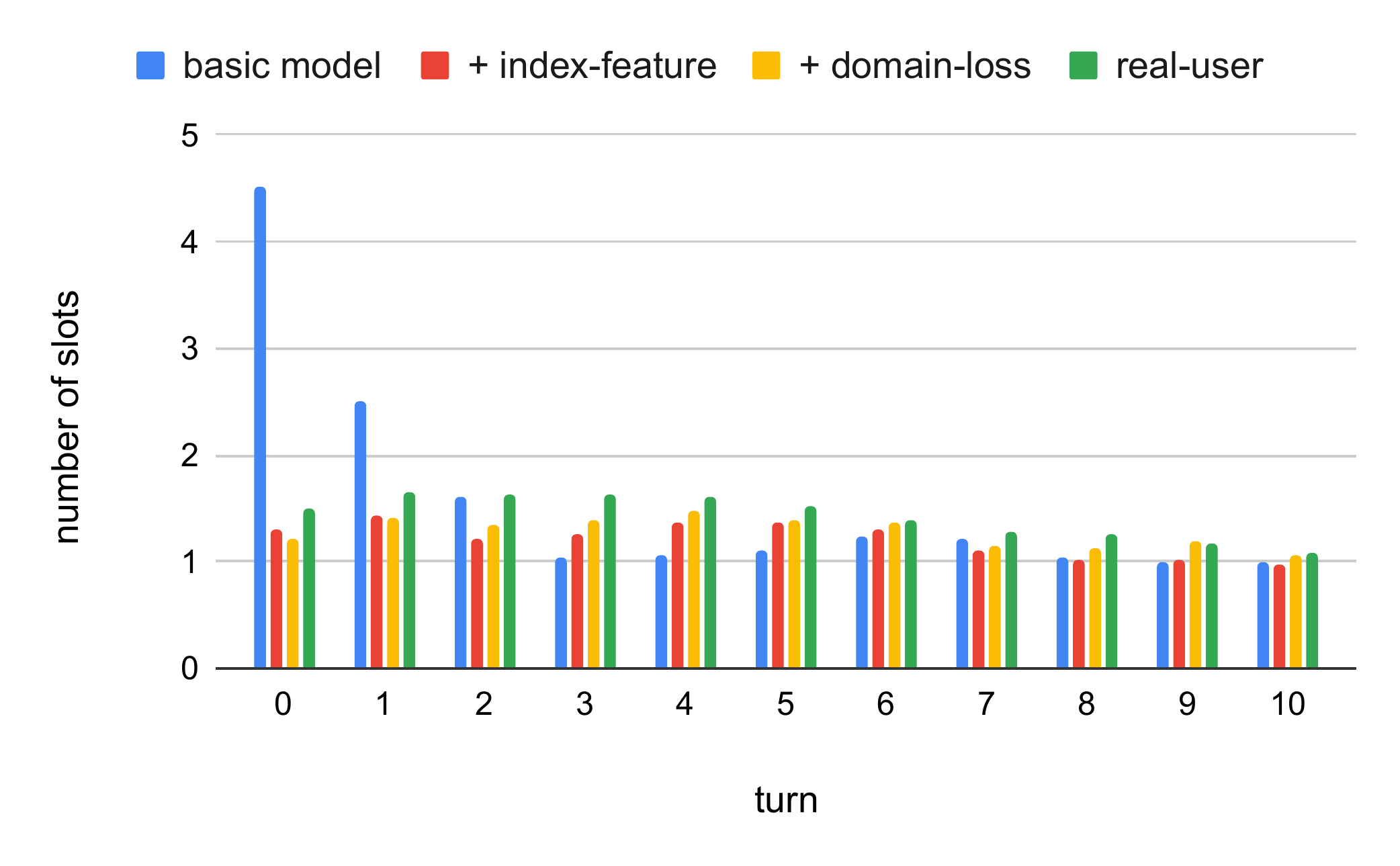}
    \caption{The average user action length per turn when interacting with the rule-based dialogue system. The average action length of real users in the corpus is also presented.}
    \label{fig:slot}
\end{figure}

The average number of slots mentioned by TUS in each turn when interacting with the rule-based dialogue system is shown in Fig.~\ref{fig:slot}. When the index feature $v_{index}$ and the domain loss $loss_{domain}$ are added, TUS can deal with the over-generation problem and behave more similarly to what is observed in the corpus.

\section{Success Rates of Leave-one-domain-out Training}
\begin{figure}[htbp]
    \centering
    \includegraphics[width=7cm]{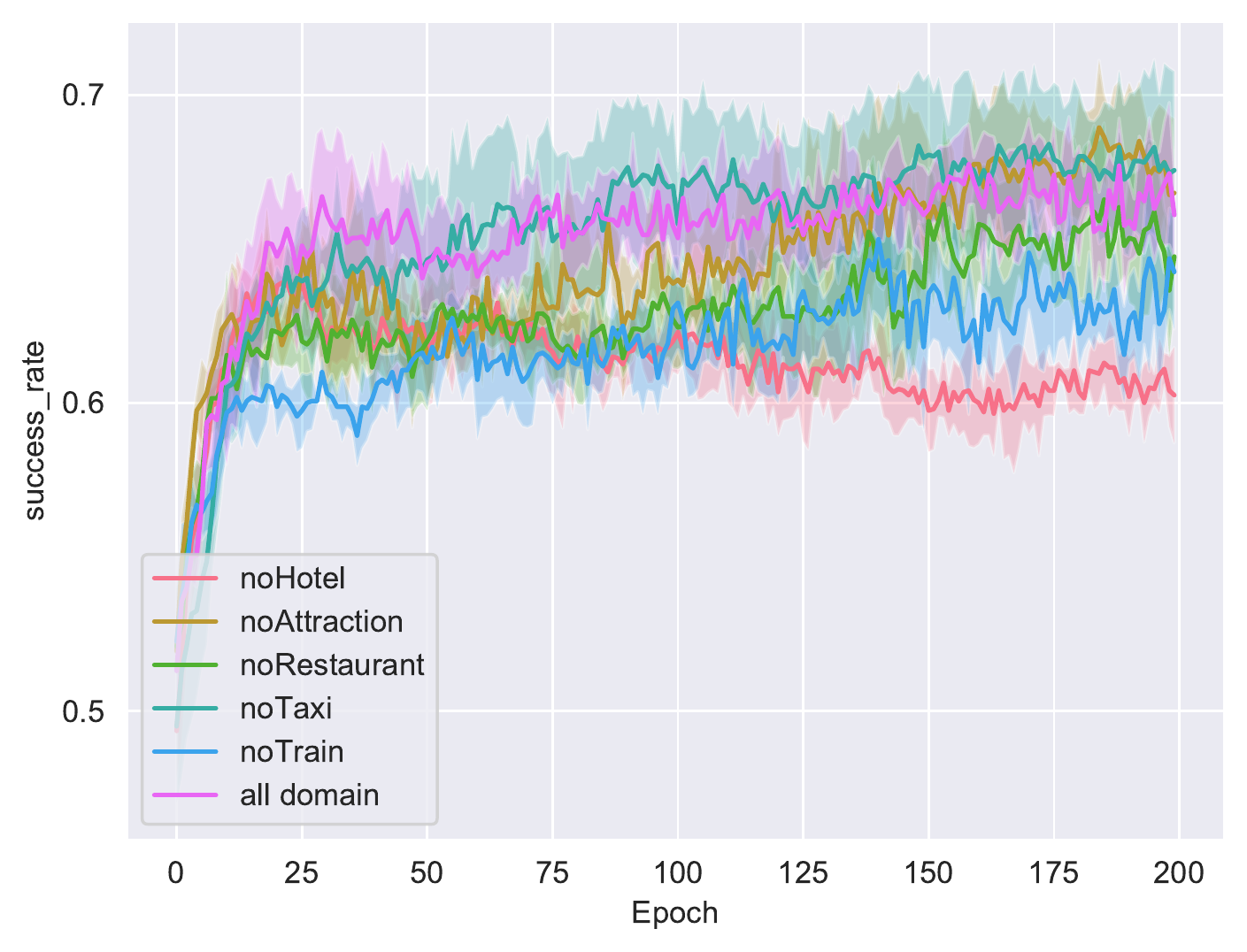}
    \caption{The success rates of dialogue policies trained with leave-one-domain-out TUSs during training, when evaluated on TUS.}
    \label{fig:domain}
\end{figure}

The training success rates of dialogue policies trained with leave-one-domain-out TUSs, which are evaluated on TUS, are shown in Fig.~\ref{fig:domain}.
In comparison to the full TUS, the leave-one-domain-out TUSs are more unstable, but they can achieve a comparable success rate at the end.

\begin{figure*}[t]
    \centering
    \includegraphics[width=16cm]{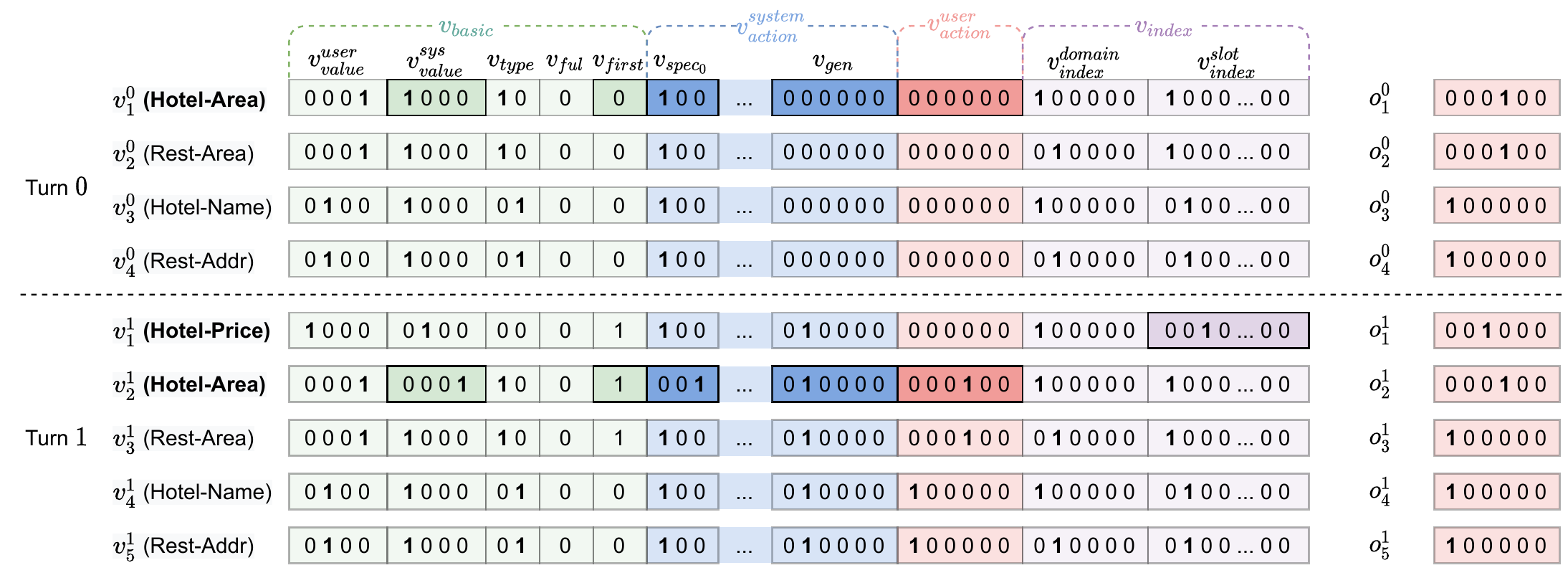}
    \caption{The input and feature representation according to Fig.~\ref{fig:task-ex}. $v_{CLS}$ and $v_{sep}$ are ignored in this graph.}
    \label{fig:feat_example}
\end{figure*}

\section{An example for the input feature representation}
\label{ap:example}

The list of input feature vectors and output sequence are presented on Fig.~\ref{fig:feat_example} based on Fig.~\ref{fig:task-ex}. 

For turn $0$, $V^0$ only includes 4 vectors from the user goal. For turn $1$, the system mentions slot \texttt{Hotel-Price}, which is not in the user goal, so the feature vector of slot \texttt{Hotel-Price} is inserted into $V^1$, where the 1-st dimension of $v^{domain}_{slot}$ is 1 because domain \textit{Hotel} is the first domain in this conversation and the 3-rd dimension of $v^{slot}_{index}$ is 1 because it is the third slot in domain \textit{Hotel}.

In comparison between the feature vectors of slot \texttt{Hotel-Area} in turn 0, $v^0_1$, and turn 1, $v^0_1$, the $v^{sys}_{value}$ and $v_{spec_0}$ are different because of the system's domain-specific action \texttt{Recom(Hotel-Area=South)}. The system also mentioned a general action, \texttt{general-reqmore()}, thus $v_{gen}$ is changed. In addition, this slot is first mentioned at turn 0, so $v_{first}$ is changed from 0 to 1. Similarly, $v^{user}_{action}$ is also modified according to the user action.
On the other hand, $v^{user}_{value}$ is the same because the user does not update its goal, $v_{type}$ is not changed because the slot is still a constraint, and $v_{ful}$ is 0 because it has not been fulfilled yet. $v_{index}^{domain}$ and $v_{index}^{slot}$ are also the same through the whole conversation.

\section{Example Dialogue Generated by TUS}
\begin{figure}[h]
    \centering
    \includegraphics[width=7cm]{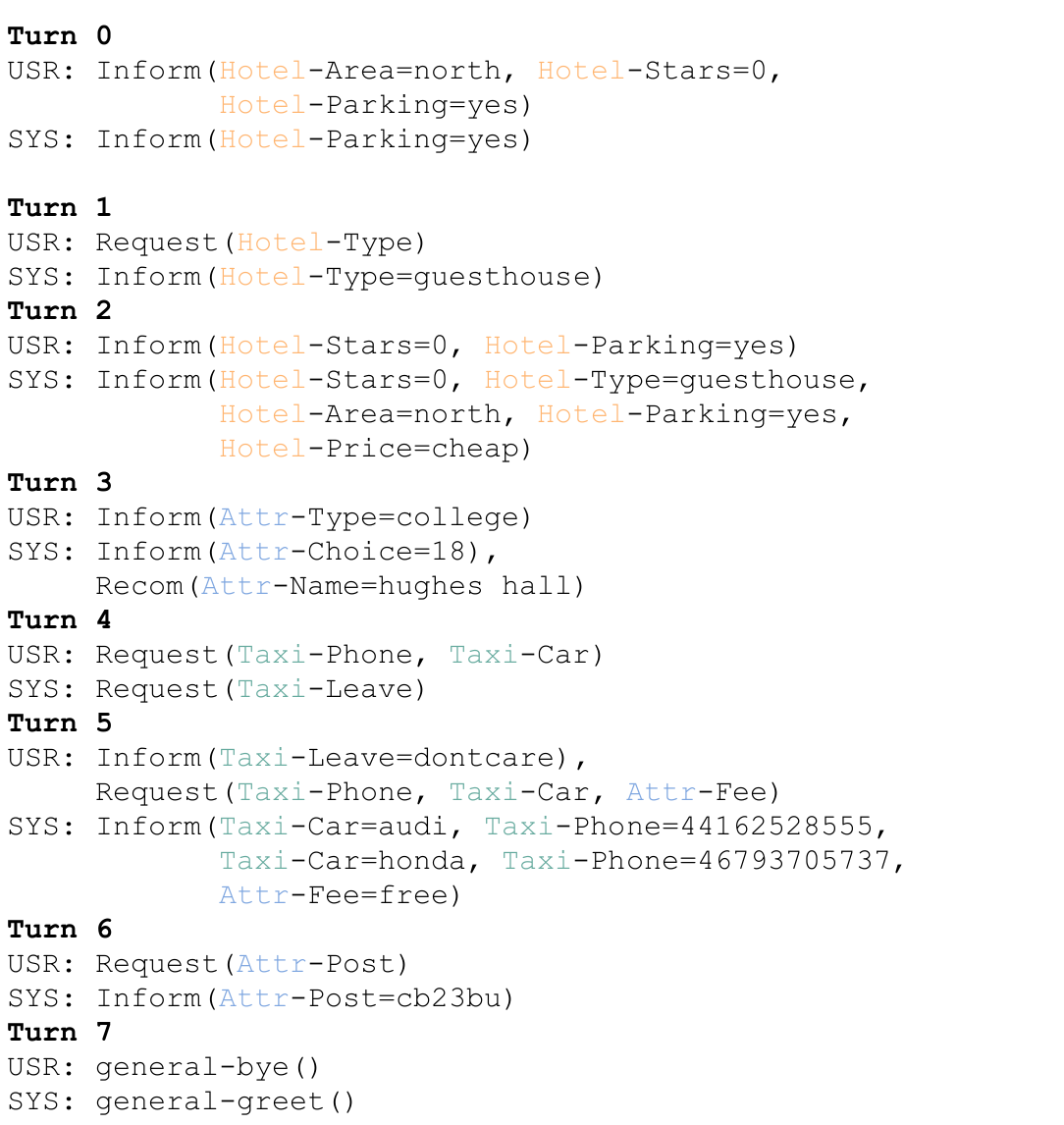}
    \caption{A dialogue generated by TUS when interacting with the rule-based policy.}
    \label{fig:large_example}
\end{figure}

An example dialogue with a multi-domain user goal is shown in Fig.~\ref{fig:large_example}. It shows that TUS is able to switch between different domains (from turn 2 to 6), respond to the system's requests, and generate multi-domain actions (in turn 5).

\end{document}